\documentclass[conference]{IEEEtran}
\IEEEoverridecommandlockouts
\usepackage{cite}
\usepackage{amsmath,amssymb,amsfonts}
\usepackage{algorithm}
\usepackage{algpseudocode}
\usepackage{graphicx}
\usepackage{textcomp}
\usepackage{xcolor}
\def\BibTeX{{\rm B\kern-.05em{\sc i\kern-.025em b}\kern-.08em
T\kern-.1667em\lower.7ex\hbox{E}\kern-.125emX}}
\usepackage{caption}
\makeatletter
\newcommand{\linebreakand}{%
  \end{@IEEEauthorhalign}
  \hfill\mbox{}\par
  \mbox{}\hfill\begin{@IEEEauthorhalign}%
}
\makeatother

\usepackage{hyperref}   
\usepackage{xurl}       

\begin{document}
\title{AI-Enhanced Radiation Classification in ROS\\
{\footnotesize \textsuperscript{}}
\thanks{$\ddagger$ Extreme Robotics Lab, University of Birmingham, Elms Rd, Birmingham, West Midlands B15 2SE, UK.}
\thanks{This work was supported in part by Birmingham Centre for Nuclear Education \& Research (BCNER).}
}
\author{

\IEEEauthorblockN{Youngseo Lee$^{\dagger\ddagger}$}
\IEEEauthorblockA{\textit{Department of Information System} \\
\textit{Hanyang University}\\
Seoul, Republic of Korea, 04763 \\
leenys@hanyang.ac.kr}

\and

\IEEEauthorblockN{Cesar Alan Contreras$^{\ddagger}$}
\IEEEauthorblockA{\textit{School of Metallurgy and Materials} \\
\textit{University of Birmingham}\\
Birmingham, UK, B15 2TT \\
cac214@bham.ac.uk}

\and

\IEEEauthorblockN{Tianshu Ruan$^{\ddagger}$}
\IEEEauthorblockA{\textit{School of Metallurgy and Materials} \\
\textit{University of Birmingham}\\
Birmingham, UK, B15 2TT}

\linebreakand

\IEEEauthorblockN{Manolis Chiou$^{\ddagger}$}
\IEEEauthorblockA{\textit{School of Metallurgy and Materials} \\
\textit{University of Birmingham}\\
Birmingham, UK, B15 2TT \\
m.chiou@bham.ac.uk}

\and

\IEEEauthorblockN{Rustam Stolkin$^{\ddagger}$}
\IEEEauthorblockA{\textit{School of Metallurgy and Materials} \\
\textit{University of Birmingham}\\
Birmingham, UK, B15 2TT \\
r.stolkin@bham.ac.uk}
}

\maketitle

\begin{abstract}
This study presents the development of two Robot Operating System (ROS) nodes with a primary focus on differentiating seven radioactive isotopes. The first node, referred to as the model training node, uses a Gradient Boosting classifier to identify and classify these isotopes. The classifier achieved a reported overall accuracy of 95\% on the held-out test dataset. The second node, known as the prediction node, provides classification results when users input data in CSV format. For the practical demonstration, the C12137 radiation detector module was mounted on the Jackal UGV, and an operator approached different radiation sources to collect data. To support future dataset expansion, we also developed a method for converting gamma-spectrum graph images into CSV files. In a separate robotic evaluation across three source locations, 30 of 42 trials were correctly classified (71.43\%). Am-241, Co-60, Cs-137, Na-22, and Ti-44 were correctly classified in all six trials each, whereas Ba-133 and Eu-152 were incorrectly classified in all six trials each. These results show lower accuracy in robotic trials than in the held-out test set.

\end{abstract}

\section{Introduction} \label{Introduction}
Detecting and classifying radioactive isotopes is important in nuclear decommissioning and robotics because isotope identification can inform source handling and operational decisions. Although recent work has demonstrated that training spectra can be generated by combining simulated isotope spectra with measured background spectra, developing labeled datasets that represent a broader range of isotopes and environmental conditions remains challenging \cite{a1}.

In this study, we use a self-collected dataset comprising measured spectra from seven radioactive isotopes to investigate isotope classification in a robotic radiation-detection setting. We developed two Robot Operating System (ROS) nodes: a model-training node using a Gradient Boosting classifier and a prediction node that processes spectrum data in CSV format and provides isotope classifications. We also developed a method for converting gamma-spectrum graph images into CSV files to support the incorporation of additional spectral data in future work.

The remainder of the paper is organized as follows. Section \ref{Literature} reviews previous research and provides context for our study. In Section \ref{Methodology}, we describe data collection, machine-learning methods, ROS-node development, gamma-spectrum graph-image conversion, and robotic testing. Section \ref{Results} presents the classifier comparison and robotic-evaluation results. Finally, Section \ref{Conclusion} summarizes our findings and suggests directions for future work.

\section{Literature Review} \label{Literature}
The differentiation of radioactive isotopes is an important aspect of nuclear decommissioning, a necessary process for ensuring nuclear safety and environmental sustainability \cite{a4}. Machine-learning methods have been applied to identify and classify radionuclides from gamma-ray spectra \cite{b1}. 

\subsection{Synthetic Datasets and Radiation Analysis}
In the domain of machine learning model training, the creation of accurate synthetic datasets is crucial. A. C. Jones et al. \cite{a5} evaluated spectroscopic anomaly-detection and isotope-identification algorithms using a synthetic urban search dataset. Similarly, L. Lee-Brewin et al. \cite{a1} generated training spectra by combining simulated isotope spectra with measured background spectra from the SIGMA dataset. They evaluated their model using both simulated spectra and real spectra from SIGMA. Together, these studies illustrate the use of synthetic gamma-ray data in isotope-identification research.

\subsection{Data Conversion Techniques Relevant to Nuclear Applications}
Recent work on extracting information from figures and documents has produced methods that may be relevant to gamma-spectrum digitization. Research by Gemelli et al. \cite{a7}, which applies graph neural networks to table extraction from PDF documents, shows progress in transforming document content into more accessible formats, enhancing data analysis and application usability. The application of the metaDigitise R package for extracting descriptive statistics from images of common plot types is another noteworthy development, contributing to improved data analysis \cite{a8}. An OpenCV-based computer-vision system for character extraction from images provides another example of visual information extraction \cite{a9}. Additionally, GFTE addresses financial-table extraction from digital files, offering a related example of structured data extraction \cite{a10}. The method developed by Hamlin for extracting data points from vector graphics to analyze electrical-resistance data is also relevant to graph digitization \cite{a11}. Together, these studies illustrate a range of approaches to extracting and digitizing information from figures and documents, providing relevant context for the gamma-spectrum graph-image conversion method investigated in this study.

\subsection{Artificial Intelligence in Isotope Identification}
Machine learning methods have been applied to radioactive-isotope identification. J. Stinnett's research \cite{a12} developed a peak-based Bayesian classifier for automated isotope identification using low-resolution gamma-ray spectra. In a related nuclear application, Braden Burt et al. \cite{a13} applied machine learning and deep learning methods to magnetic-field sensor data to classify the High Flux Isotope Reactor's ON/OFF operational status; the DNN achieved 95.70\% balanced accuracy for cycle 488. 

Complementing these developments, studies have stressed the capabilities of artificial neural networks in isotope identification. M. Park et al. \cite{a14} introduced a deep learning-based radionuclide identification algorithm, while L. Lee-Brewin et al. \cite{a15} developed a convolutional neural network for multi-isotope detection. M. J. Neuer and C. Henke \cite{a16} proposed an autoencoder-based method for identifying marker isotopes in special nuclear materials. However, developing large and representative gamma-ray spectral datasets remains challenging for isotope identification \cite{a12,a15}. Our study contributes a dataset comprising spectra from seven isotopes and a method for converting gamma-spectrum graph images into CSV files, which may support future dataset expansion.

\subsection{Advancements in ROS for Nuclear Applications}
Integrating advanced technologies with the ROS supports practical mobile-robot applications. H. Ahmad et al.'s research \cite{a17} shows the use of ROS for environment modeling and mobile-robot position estimation under uncertain conditions. Separately, Ye et al. \cite{a18} developed a neural network combining Convolutional Neural Networks (CNN) and Long Short-Term Memory (LSTM) networks to generate robot actions based on current and historical scenes and robot movements.This historical context enables the network to use past observations when selecting actions to obtain a better view of an object for detection. 

The convergence of machine learning, neural networks, data conversion techniques, and ROS integration may create opportunities for further research on radioactive isotope differentiation in nuclear decommissioning and safety. Future studies could investigate potential applications in other areas requiring precise identification and handling of hazardous materials, such as healthcare and the chemical industry.

\section{Methodology} \label{Methodology}
Our study integrates three key components to enhance isotope classification and real-time detection in nuclear environments. Initially, a detailed dataset of seven radioactive isotopes was compiled using the C12137 radiation detector module. We then developed a method to convert gamma spectrum graphs into CSV files to support future dataset expansion, utilizing image preprocessing and binarization. In terms of machine learning, the model\_training\_node employed a Gradient Boosting classifier for training on spectral data, while the prediction\_node processed input spectrum data for isotope identification. Finally, our methodology was evaluated through robotic testing, where the Jackal UGV equipped with the C12137 radiation detector module collected and processed data, which were then processed using the \textit{prediction node}.

\subsection{Data Collection}
In our study, we collected a dataset comprising seven distinct radioactive isotopes for isotope classification in a robotic radiation-detection setting. Twenty-five spectra were collected for each isotope at varying distances. Additional information in the dataset can be found in this section.

i) Dataset Overview: Each entry in the dataset includes energy and corresponding count data for various radioactive isotopes, each displaying distinct radiation activities. The energy is measured in kiloelectronvolts (keV), serving as the standard unit for quantifying the energy of small particles or atoms. The dataset encompasses seven radiation sources: 
\begin{itemize}
  \item Na-22 (Sodium-22): Utilized in radiation measurements and experiments.
  \item Ti-44 (Titanium-44): Employed in nuclear physics experiments and astrophysics research.
  \item Co-60 (Cobalt-60): Used in radiation therapy and radiation therapy equipment.
  \item Ba-133 (Barium-133): Employed in medical imaging and radiation therapy.
  \item Cs-137 (Cesium-137): Used for radiation contamination detection and nuclear test tracking.
  \item Eu-152 (Europium-152): Utilized in radiation standards and medical imaging.
  \item Am-241 (Americium-241): Used in smoke detectors and moisture gauges.
\end{itemize}
Radioactive materials find applications in many fields like industry, medicine, agriculture, research, education, and military operations. The International Basic Safety Standards establish a globally unified framework for the safe use of ionizing radiation sources. The concept of a 'dangerous' source, assessed through 'D values,' signifies a source capable of causing exposure leading to severe deterministic effects if uncontrolled. While 'D' values for certain isotopes like Na-22, Ti-44, Ba-133, and Eu-152 are pending confirmation, the IAEA has provided values for Co-60, Cs-137, and Am-241. Notably, Cs-137 holds the highest 1000 x D value at 1.0E+02 TBq, indicating a 'dangerous' activity level, emphasizing its greater risk compared to other substances. The 1000 x D values for Am-241 and Co-60 are 6.0E+01 TBq and 3.0E+01 TBq, respectively, signifying varying levels of concern \cite{a19}. 

ii) Data Acquisition and Setup: Data were collected using the C12137 radiation detector module, which supports gamma-ray detection over an energy range of 30 keV to 2 MeV. This module contains a scintillator coupled to an MPPC (multi-pixel photon counter). The scintillator converts gamma rays into visible light, which the MPPC detects \cite{a20}. The experiment was conducted on an x86\_64 computer running Ubuntu 20.04.5 LTS. The computer was equipped with an Intel(R) Core(TM) i7-8086K CPU (six cores, 12 threads, and a base frequency of 4.00 GHz) and 32 GB of RAM. Radiation sources were placed at predetermined distances from the detector. The sampling rate was set at 5 Hz to acquire 300 samples for each saved spectrum, giving an acquisition time of 60s per spectrum. The collected data were recorded in a CSV file with two columns, 'Energy [keV]' and 'Count,' containing 4096 energy-count pairs.

iii) Distance-Dependent Variation in Spectral Peak Counts: In our measurements, spectral peak counts generally decreased as the source-to-detector distance increased. Five spectra were collected at each of the following distances: 1 cm, 3 cm, 5 cm, 7 cm, and 9 cm, yielding 25 spectra per isotope. However, because the 9 cm spectra showed weak or inaccurate peaks, all spectra collected at 9 cm were excluded from training the isotope-prediction model.

\begin{figure}[h]
  \centering
  \includegraphics[width=85mm]{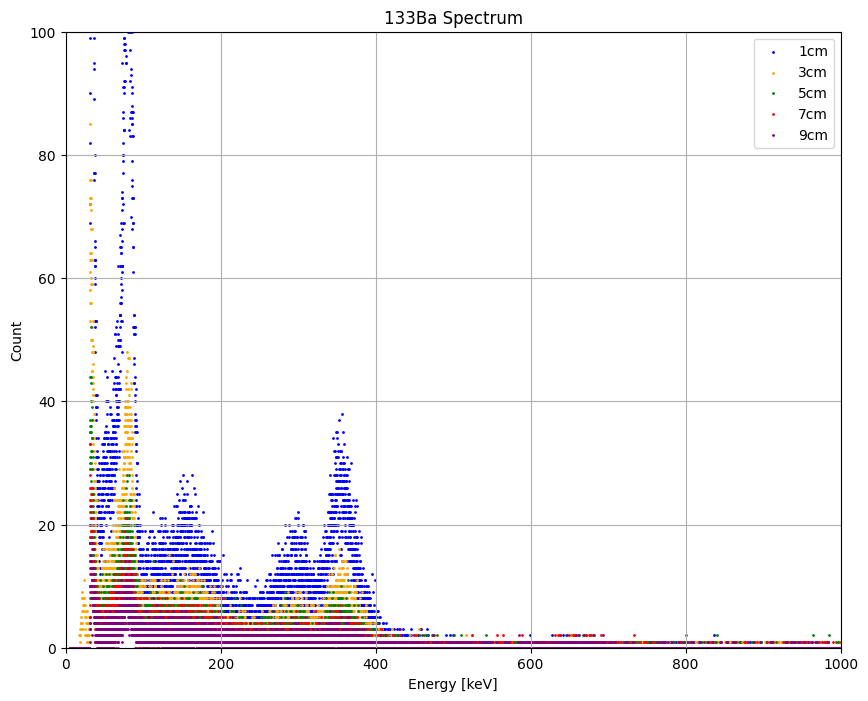}
  \caption{Ba-133 gamma-ray spectra measured at different source-to-detector distances}
  \label{fig:ba133-distance-variation}
\end{figure}

Figure~\ref{fig:ba133-distance-variation} shows the measured Ba-133 gamma-ray spectra, with counts plotted against energy (keV). The colors indicate measurements taken at different source-to-detector distances. In these measurements, spectra collected at shorter distances generally show higher counts.

\subsection{Conversion of Gamma Spectrum Graph Images into CSVs}
In our research, we have developed a method for the integration of additional isotopes by transforming gamma spectrum graph images into CSV files. This approach empowers researchers to easily extract isotope values from gamma spectrum images and augment the existing dataset with new data. The method provides flexibility for future expansion and refinement of the dataset. This involves the following key steps: 

i) Image Preprocessing: The initial step involved loading the original gamma spectrum graph image and converting it to grayscale. Subsequently, a targeted region of interest was identified and cropped to isolate and concentrate on the pertinent data for further analysis. 
\begin{figure}[h] 
  \centering
  \includegraphics[width=85mm, scale=1]{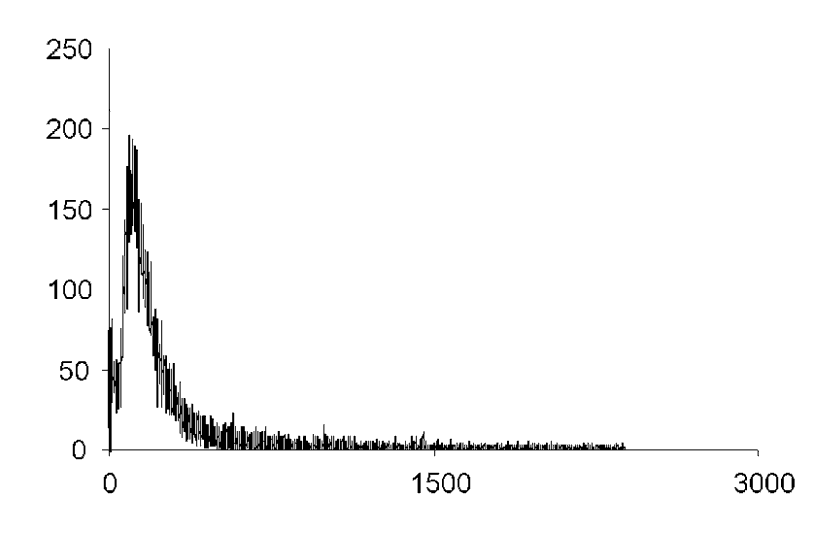}
  \caption{Graph image of Sr-90}
  \label{Graph image of Sr-90}
\end{figure}

\begin{figure}[h] 
  \centering
  \includegraphics[width=85mm, scale=1]{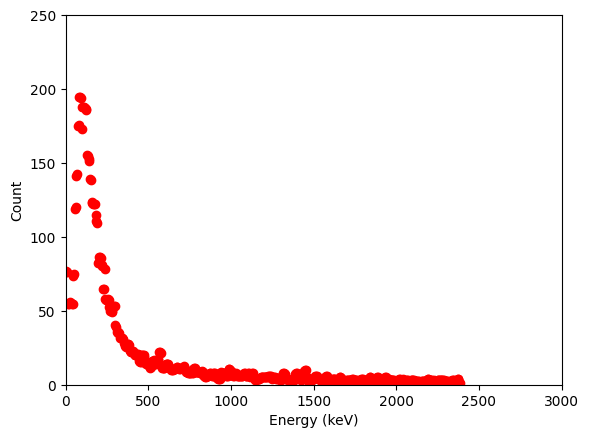}
  \caption{Sr-90 gamma spectrum graph, transformed from the newly created CSV file}
  \label{fig:gammaspectrumgraph}
\end{figure}

In this study, the gamma spectrum graph image of Strontium-90 (Sr-90) served as a test subject. Sr-90 was chosen due to its pivotal role in the field of radiation as a byproduct of nuclear fission, commonly present in nuclear waste and fallout. Its emission of high-energy beta radiation renders Sr-90 crucial for the investigation of radioactive contamination, environmental monitoring, and the assessment of potential health risks associated with nuclear activities. 

ii) Image Scaling and Binarization Process: We defined the X and Y scales of the image, representing how many kilo electronvolts (keV) or counts each pixel corresponds to. The blurred image underwent thresholding to create a binary mask, where pixels with a brightness level above 250 were set to white (255), and those below were set to black. For each X-coordinate, we identified the topmost Y-coordinate of black pixels. If black pixels were present, their corresponding Y-coordinate was recorded; otherwise, the Y-coordinate was set to 0. Considering that the Y-axis increases from top to bottom in image coordinates, we inverted the Y-coordinates.

iii) CSV File Creation and Evaluation: After extracting key coordinates, they were compiled into a CSV file, categorizing them under 'Energy (keV)' and 'Count' columns. This file forms the foundation for further analysis. To assess this conversion, we generated a plot (Figure \ref{fig:gammaspectrumgraph}) depicting the relationship between energy levels and counts. This visual representation, derived from the Sr-90 gamma spectrum graph and reconstructed from the CSV file, provides a qualitative check of our transformation. It confirms the precision of our method and the significance of converting complex graphical data into a more analyzable and accessible format.

\subsection{Integration of ROS Python Nodes with Robotics for Isotope Detection}
Our research involved the development of two specialized ROS Python nodes: the model\_training\_node and the prediction\_node. These nodes were designed to use the radioactive isotope dataset and machine learning methods for isotope classification and identification. 

i) Model Training Node: We employed the ‘find\_peaks’ function from the ‘scipy.signal’ module to extract significant peaks from spectral data, focusing on peak locations and corresponding counts. The function identifies local maxima and selects peaks according to criteria such as height, width, and distance. Following peak extraction, these features were organized into arrays, maintaining uniform length through padding. This standardization ensures consistent and efficient data processing, essential for machine learning applications. 

The dataset, containing seven isotopes with 15 retained peaks per CSV file, was used to train a Gradient Boosting classifier. On the held-out test set, the classifier achieved a reported overall accuracy of 95\%. In-depth evaluations of the model, particularly comparing the accuracy of the Gradient Boosting classifier with other classifiers, are detailed in Section \ref{Results}. 

The prediction node employs seven isotope-specific classifiers. Each classifier was evaluated using test spectra from all seven isotope types. The output score of each classifier is compared with a 60\% threshold, and the system reports no match when none exceeds the threshold. This evaluation covered the seven isotopes in the dataset; performance on background-only measurements and isotopes outside the dataset was not evaluated.

Subsequently, the dataset was divided into training and testing subsets to facilitate model evaluation. 30\% of the data, was allocated for the test set, while the remaining 70\%, was used as the training set. After training, the model’s precision, recall, and F1-score were analyzed using the test dataset, yielding results for each isotope’s classifier. Each isotope was individually tailored and preserved as a different model, resulting in a total of seven models. We saved the trained models using 'joblib', enabling their future application without retraining.

ii) Prediction Node: The prediction\_node, implemented in Python and integrated with ROS, provides isotope classifications from spectral data. This node begins by extracting significant features from a provided spectrum file, a process akin to that in the model\_training\_node. The feature extraction function reads spectrum data from a CSV file and utilizes the ‘find\_peaks’ method from the ‘scipy.signal’ module to identify peaks, recording their energy and count values. 

Subsequently, the top 15 peaks, determined by their count values, are selected and transformed into a single array, aligning with the trained model's input requirements. This array is zero-padded as necessary to achieve a uniform length of 30 features. The feature array is then reshaped to be compatible with the Gradient Boosting classifier model. The prediction\_node employs seven previously trained models, loaded using ‘joblib’, to execute the prediction. 

Upon model activation, the prediction node evaluates the spectral data against each specific isotope, applying a 60\% confidence threshold. This threshold ensures that an isotope is identified only if the probability exceeds this limit. We categorized output scores into three distinct levels—“possible match” for scores from 60\% to below 80\%, “high-score match” for scores from 80\% to below 95\%, and “very-high-score match” for scores of 95\% or higher.

In scenarios where multiple isotopes exhibit probabilities scores above this threshold, the system suggests an ambiguity in the prediction, especially if the actual sample contains only a single isotope. Conversely, if none of the isotopes go over a confidence threshold exceeds the score threshold, the system communicates that no matching isotope has been found. The models then analyze the input features to ascertain the most likely isotope presence, formatting in a human-readable message that not only identifies the isotope but also recommends appropriate action. The prediction result is formatted as a human-readable message that identifies the isotope and recommends an appropriate action. This advisement is derived from a predefined isotope actions dictionary, aligning the prediction with actionable insights. Furthermore, this information is broadcast over a ROS topic, enhancing the utility of the prediction\_node within the ROS ecosystem.

iii) Robotic Testing: 
Our study features the Jackal UGV's utilization alongside the C12137 radiation detector module for mapping and detecting radiation. The robot generates a map for tracking its location in real-time. Upon approaching a radiation source, it displays a warning, and then collects and analyzes raw data, classifying it into specific isotopes. This informs future reactions from both robots and humans.

In this study, we employed the Jackal UGV for mapping and navigating potentially hazardous environments due to radioactive isotopes. The robot creates a graphical map while saving metadata including map size, resolution, and initial positioning, enabling real-time tracking and operator-guided navigation. Through the Adaptive Monte Carlo Localization (AMCL) and visualization tools, the robot's location can be monitored. When the robot approaches a radiation source, a warning message will be displayed in the console. As a result, users can determine the approximate location of the radiation source. This process enables future actions based on the specific isotope identified, utilizing Human Robot Interaction (HRI).

For the trials, three distinct locations for isotope placement were selected: center, left, and right. The source was placed on top of the bricks. Initially, the environment was set up by situating one of the seven isotopes at a predetermined spot. Following this, the C12137 radiation detector module was equipped on a robot, allowing the robot to detect gamma radiation emitted by the isotopes. The Jackal UGV was then deployed, with its starting point at the map's center. As an operator drives the robot towards the isotope, a warning signal was triggered to alert the user of its proximity to the radiation source. Each of the chosen locations was visited twice, resulting in a total of 6 trials per isotope and cumulatively, 42 experiments across the study.

Three operators alternated in controlling the robot to ensure diversity and fairness in the isotope detection trials across three locations. Two data collectors alternated recording trial number, location, isotope values, time, distance, predicted values, and the name of the saved CSV file. The time to reach a location ranged from 8 to 17 seconds across operators. The distance between the robot and the radiation source varied from 2 to 8.3 cm, with a minimum of 2 cm due to the height difference between the robot and the source, encompassing measurements in all directions.

\begin{figure}[h] 
  \centering
  \includegraphics[width=85mm, scale=1]{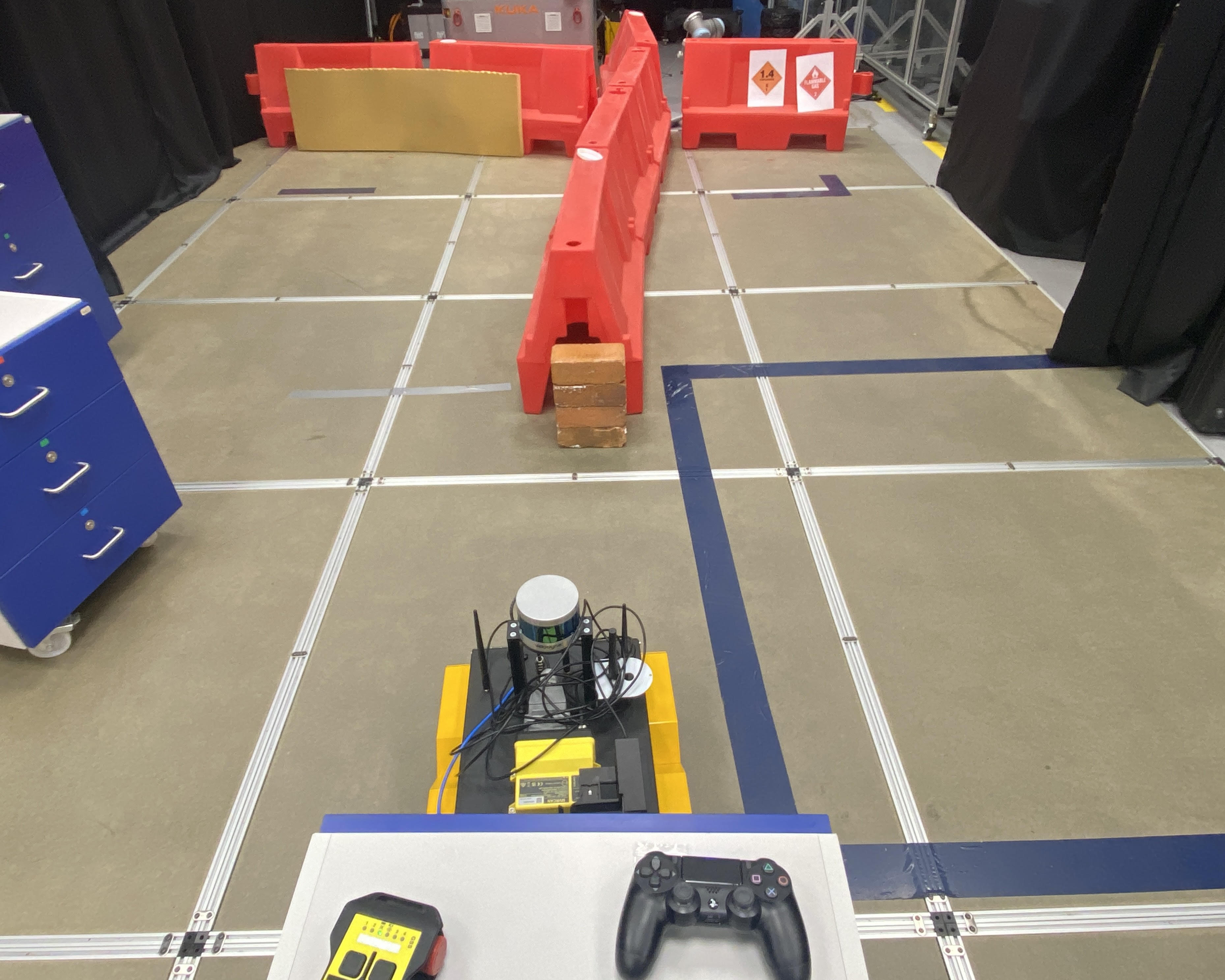}
  \caption{The environment image, representing the initial location of the Jackal UGV and 3 chosen locations}
  \label{The environment image, representing the initial location of the Jackal UGV and 3 chosen locations}
\end{figure} 

When driving and positioning the robot very close to the isotope, it remained stationary for a minute to collect raw data. This data collection process mirrored the conditions under which the initial dataset was compiled. The radiation detector automatically recorded the collected data in a CSV file, formatted with two columns: 'Energy [keV]' and ' counts,' totaling 4096 energy channels. There were instances when the radiation detector malfunctioned, resulting in counts of zero. In such cases, the procedure was repeated to ensure accuracy.

Following the collection of spectral data, we utilized the prediction node to classify the isotope. This node processed the CSV file, applying the previously trained machine-learning model to predict the type of isotope according to its spectral features. Based on the prediction\_node's output, different protocols or actions were assigned to the robot. These responses were tailored according to the identified isotope, to influence robotic behavior in real-time. With a closed-loop system where data collection, analysis, and action are seamlessly connected. The operator controlled the robot during the trials, and the prediction node published the classification result and associated action recommendation over a ROS topic. This research overall demonstrates an enhancement in robotics and sensor technology, improving safety through real-time mapping, detection, isotope classification, and Human-Robot Interaction (HRI). Overall, this study demonstrates the integration of radiation detection, isotope classification, and ROS-based communication in an operator-guided robotic test.

Although this research stresses an improvement in the nuclear field, it inevitably contains two main limitations. First, the Gradient Boosting model used in the prediction node distinguishes isotopes based on the number of spectral peaks and counts, making it challenging to achieve accurate detection in environments where multiple types of isotopes are present. Additionally, for accurate isotope identification, the robot needs to approach the radiation source, which limits its operational range and efficiency. This reliance on the operator's instinct to decide when to move closer introduces an additional constraint, underscoring the necessity for enhancing the system's autonomy to mitigate human judgment errors and optimize performance. 

\section{Experimental Results} \label{Results}
In the AI-enhanced radiation classification study, precision, recall, and F-score are critical metrics for evaluating the performance of the classification model. True Positives, denoted as  \textit{TP}, represent instances where the model accurately identifies the presence of a condition, such as correctly detecting a specific isotope. False Positives, represented as \textit{FP}, occur when the model incorrectly identifies a condition. False Negatives, denoted as  \textit{FN}, are cases where the model fails to detect the condition, such as missing the presence of an isotope. 

Precision, calculated as
\[ \frac{TP}{TP + FP} \],
measures the proportion of true positive predictions in all positive predictions made, essential for understanding the accuracy of isotope classification. Recall, 
\[ \frac{TP}{TP + FN} \],
evaluates how effectively the model identifies all relevant instances of a specific isotope. The F-score, the harmonic mean of precision and recall, is represented as 
\[ 2 \times \frac{\text{Precision} \times \text{Recall}}{\text{Precision} + \text{Recall}} \],
balances both metrics, providing a comprehensive view of the model's performance in distinguishing between various radioactive isotopes within a robotic system. Finally, support represents the number of occurrences of each class in the actual dataset, indicating the count of true instances for each class in our dataset \cite{a21}. 

\begin{table}[ht]
\caption{Classification Performance on the Held-Out Spectral Test Set} 
\centering 
\begin{tabular}{c c c c c c} 
\hline 
 &\ Precision &\ Recall &\ F1-score  &\ Support\\ 
\hline 
Am-241 & 1.00 & 1.00 & 1.00 & 7 \\
Ba-133 & 1.00 & 1.00 & 1.00 & 5 \\
Co-60 & 1.00 & 0.67 & 0.80 & 6 \\
Cs-137 & 1.00 & 1.00 & 1.00 & 5 \\
Eu-152 & 0.80 & 1.00 & 0.89 & 8 \\
Na-22 & 1.00 & 1.00 & 1.00 & 6 \\
Ti-44 & 1.00 & 1.00 & 1.00 & 5 \\
\hline 
Accuracy &  &  & 0.95 & 42 \\
Macro Avg & 0.97 & 0.95 & 0.96 & 42 \\
Weighted Avg & 0.96 & 0.95 & 0.95 & 42 \\
\hline 
\end{tabular}
\label{table:weight_one} 
\end{table}

The overall accuracy of Table \ref{table:weight_one} is 95\% for the retained test split on the classification of various isotopes. This table was derived using the one-versus-all method. In the one-versus-one scenario, which is employed in the ROS node, the accuracy rates observed were very similar. The element Ba achieved 98\%, Co reached 93\%, and the remaining five isotopes each attained a perfect accuracy of 100\%. Accuracy measures the overall correctness of classification, macro avg represents the average F1-score across all classes, and weighted avg calculates the weighted average of the F1-score, taking class imbalances into account. 

\begin{table}[ht]
\caption{Comparison of Classification Performance Metrics for Different Classifiers} 
\centering 
\begin{tabular}{c c c c c} 
\hline 
Classifier &\ Accuracy &\ Macro Avg &\ Weighted Avg \\ 
\hline 
Gradient Boosting & 0.95 & 0.96 & 0.95 \\
Decision Tree & 0.95 & 0.96 & 0.95 \\
Support Vector Machine & 0.90 & 0.91 & 0.91 \\
Random Forest Classifier & 0.90 & 0.91 & 0.90 \\
Logistic Regression & 0.83 & 0.83 & 0.84 \\
\hline 
\end{tabular}
\label{table:weight_two} 
\end{table}

Table \ref{table:weight_two} compares five different classifiers for isotope classification based on three key performance metrics: accuracy, macro avg (average F1-score), and weighted avg. This table was also derived using the one-versus-all method. 

It shows that Gradient Boosting and Decision Tree achieved the same reported performance on the held-out test split, with an accuracy of 95\%. In a seperate evaluation of the deployed prediction-node classifiers, Gradient Boosting achieved higher accuracy than Decision Tree. The Gradient Boosting classifier achieved per-isotope accuracies of 98\% for Eu-152 and Ba-133, 95\% for Co-60, and 100\% for each of the remaining four isotopes. Gradient Boosting was therefore selected for deployment in the ROS prediction node.

\begin{table}[ht]
\caption{Isotope Detection Efficacy Across Locations} 
\centering 
\begin{tabular}{c|cc|cc|cc} 
\hline
Isotope & \multicolumn{2}{c|}{Center} & \multicolumn{2}{c|}{Left} & \multicolumn{2}{c}{Right} \\ 
\hline
Trial & 1 & 2 & 1 & 2 & 1 & 2 \\ 
\hline
Am-241 & O & O & O & O & O & O \\
Ba-133 & X & X & X & X & X & X \\
Co-60 & O & O & O & O & O & O \\
Cs-137 & O & O & O & O & O & O \\
Eu-152 & X & X & X & X & X & X \\
Na-22 & O & O & O & O & O & O \\
Ti-44 & O & O & O & O & O & O \\
\hline
\end{tabular}
\label{table:isotope_trials} 
\end{table}

Table \ref{table:isotope_trials} shows the outcomes of isotope detection trials conducted at three distinct locations (Center, Left, Right) with each isotope (Am-241, Ba-133, Co-60, Cs-137, Eu-152, Na-22, Ti-44) tested twice per location. The trials are represented as ’O’ for correct classification and ’X’ for incorrect classification after the robot was manually positioned close to the isotopes. This evaluation illustrates the detection system's consistency across various environmental settings. 

\section{Discussion and Conclusions} \label{Conclusion}
\subsection{Interpretation of Results}
The experimental results of this study provide a comprehensive evaluation of the proposed AI-enhanced radiation classification system within a Robot Operating System (ROS) framework. The classification performance metrics—precision, recall, F1-score, and support—offer a detailed insight into the model's ability to accurately identify seven key radioactive isotopes. The Gradient Boosting Classifier’s accuracy of 95\%, equal to Decision Tree and above Random Forest in Table \ref{table:weight_two}, reflects its classification performance on the held-out spectra.  

The robotic test results show that Ba-133 and Eu-152 were misclassified in all six trials each possibly due to similar energy peak positions and low count values. Data analysis revealed that while 1cm distances showed discernible peaks, any distance beyond this threshold resulted in peak counts falling below 100, complicating the separation of peaks from background noise. Despite attempts to position the radiation detector as close to the source as possible, a minimum 2cm height difference and the manual operation limited proximity. This resulted in indistinct peak data, with counts all below 100, leading to Ba not being correctly classified at 0.0\% and Eu being incorrectly identified as Ba due to similar energy peak positions and low count values. This interpretation underscores the need for enhanced detection strategies for weak isotopes in practical settings. 

\subsection{Conclusion}
This study developed and integrated advanced nodes for identifying seven key radioactive isotopes. Two nodes, a model training node and a prediction node, were built. The former, employing a Gradient Boosting classifier, showed an accuracy of 95\% on the held-out test set in isotope classification from spectral data. The latter processes spectrum data in CSV format and returns a classification result. We collected and prepared the dataset used in this study, ensuring reliability in the classification process. Additionally, a method to transform gamma spectrum graphs into CSV files was introduced allowing for the inclusion of more isotopes and enhancing our methodology's flexibility and application in diverse nuclear scenarios. This approach also addresses the challenge of isotope scarcity due to safety issues to support future dataset expansion when direct measurements are limited by safety concerns. In the robotic evaluation, 30 of 42 trials were correctly classified (71.43\%). The successful merging of these ROS nodes with robotic systems during practical testing highlights their real-world applicability integration in the tested robotic setup. Our research therefore advances nuclear decommissioning and variable autonomy robotics, offering an initial workflow for isotope detection and classification.

\bibliographystyle{IEEEtran}
\bibliography{references}

\end{document}